# A Probabilistic Approach for Learning Folksonomies from Structured Data

Anon Plangprasopchok,* Kristina Lerman,† and Lise Getoor‡

(Dated: November 15, 2010)

Learning structured representations has emerged as an important problem in many domains, including document and Web data mining, bioinformatics, and image analysis. One approach to learning complex structures is to *integrate* many smaller, incomplete and noisy structure fragments. In this work, we present an unsupervised probabilistic approach that extends affinity propagation [7] to combine the small ontological fragments into a collection of integrated, consistent, and larger folksonomies. This is a challenging task because the method must aggregate similar structures while avoiding structural inconsistencies and handling noise. We validate the approach on a real-world social media dataset, comprised of shallow personal hierarchies specified by many individual users, collected from the photosharing website Flickr. Our empirical results show that our proposed approach is able to construct deeper and denser structures, compared to an approach using only the standard affinity propagation algorithm. Additionally, the approach yields better overall integration quality than a state-of-the-art approach based on incremental relational clustering.

## I. INTRODUCTION

Learning structure from data has emerged as an important problem in many domains, for example, learning gene networks from microarray data [5] and learning the structure of probabilistic networks from knowledge bases [10]. Here we focus on learning complex structures from data that may already be explicitly structured, albeit more simply.

Learning complex structures from collections of many small, simple structures may provide insights into data that individual small structures cannot provide. To infer complex structures one needs machinery to manipulate and combine structured data. For example, in order to find communities of authors of scientific papers, one must first identify individual entities appearing among author names in co-authorship network [1], then aggregate co-authorship relations between the identified entities. To learn complex structures of a specific form, such as a *tree* or a *directed acyclic graph*, the integration method must have extra machinery to avoid structural inconsistencies, that are likely to appear when data is combined arbitrarily. The task becomes even more challenging when data comes from numerous heterogeneous sources. Such data is inherently noisy and inconsistent, and there is certainly no single, unified structure to be found that explains all the data.

One instance of such a task is learning a taxonomy from many smaller trees generated by many people: the so-called folksonomy learning task [16]. In folksonomy learning, the input, structured metadata in the form of hierarchies of conceptual terms created by individual users, is combined into a global taxonomy that reflects how a community organizes knowledge. Users who create personal hierarchies to organize content may use idiosyncratic categorization schemes [9] and naming conventions. Simply combining nodes with similar names is very likely to lead to ill-structured graphs containing loops and shortcuts (multiple paths from one node to another), rather than a tree. The folksonomy learning problem has been addressed in recent work [17] using a bottom-up approach to heuristically construct the folksonomy.

In this paper, we present a more flexible probabilistic framework for learning complex structures with a specific form from fragments of structured data by exploiting structural information. Our approach extends affinity propagation [7] to use structural information to guide the inference process to combine data *concurrently* into more complex structures with a desired form. We examine two strategies for introducing structural information into affinity propagation: through the *similarity function* and through *constraints*.

## II. LEARNING FOLKSONOMIES BY INTEGRATING STRUCTURED METADATA

We take as our motivating example user-generated structured metadata on the Social Web. We assume that groups of users share common conceptualizations of the world, which can be represented as a taxonomy or hierarchy of concepts. Figure 1(a) depicts one such common conceptualization about 'animal' and its 'bird' subconcepts shared by a group of users. When users organize the content they create, e.g., photographs on Flickr, they select some portions of the common taxonomy for categorization. We observe these categories through the shallow personal hierarchies Flickr users create. There personal hierarchies, which we refer to as *saplings*, are similar to how users organize their computer files within folders and subfolders. Figure 1(b) depicts some of the saplings specified by different users to organize their 'animal' and 'bird' images. Our ultimate goal is to infer the common conceptual hierarchy related to 'animal' from the individual saplings. One natural solution is to aggre-gate saplings shown in Figure 1(b) together into a deeper and bushier tree shown in Figure 1(a).

---
*Electronic address: anon.plangprasopchok@nectec.or.th
†Electronic address: lerman@isi.edu
‡Electronic address: getoor@cs.umd.edu

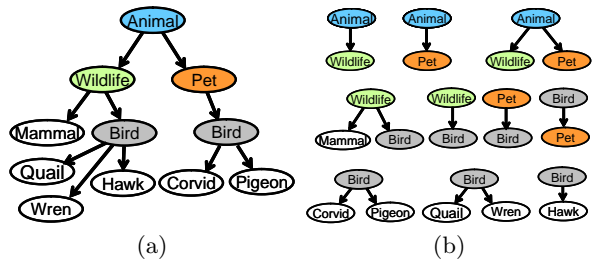

FIG. 1: Illustrative examples on (a) a commonly shared conceptual categorization (hierarchy) system; (b) personal hierarchies expressed by the users based on the conceptual categorization in (a). For illustrative purposes, nodes with similar names have similar color.

To learn a common tree by aggregating saplings, we need a strategy that measures the degree to which two sapling nodes are similar, and therefore, should be merged. Suppose that we have a very simple aggregation strategy that says two nodes are similar if they have similar names as in the prior work [16]. From Figure 1(b), we will end up with a graph containing one loop and two paths from 'animal' to 'bird', rather than the tree shown in Figure 1(a). Suppose that we can also access tags with which users annotated photos within saplings, and that photos one of the 'bird' nodes have tags like "pet" and "domestic" in common, while photos belonging to the other 'bird' node have tags like "wildlife" and "forest" in common. A cleverer similarity function that, in addition to node names, takes tag statistics within a node into consideration, should split 'bird' nodes into two different groups: 'pet bird' and 'wild bird', which are put under 'pet' and 'wildlife' nodes respectively.

The similarity function plays a crucial role in sapling integration process, and a sophisticated enough similarity function that can differentiate node senses may potentially correctly integrate the final tree. However, finding and tuning such function is very difficult. Moreover, data is often inconsistent, noisy and incomplete, especially on the Social Web, where data is generated by many different users.

One possible way to tackle this challenge is to use a simple similarity function and incorporate constraints during the merging process. Intuitively, we would not consider merging the 'bird' node under 'pet' with the one under 'wildlife' because it will result in multiple paths from 'animal'. Specifically, we can impose constraints that will prevent two nodes from being merged if (1) this will lead to links from different parent concepts or (2) this will lead to an incoming link to the root node of a tree. These constraints guarantee that there is, at most, a single path from one node to another.

## A. Personal Hierarchies in Flickr

Structured data in the form of shallow hierarchies is ubiquitous on the Social Web. The social bookmarking site *Delicious* allows users to bundle related tags together. On *Flickr*, users can group related photos into *sets* and then group related sets in *collections*. Some users create multi-level hierarchies containing collections of collections, etc., but the vast majority of users who use collections create shallow hierarchies, consisting of collections and their constituent sets. These personal hierarchies generally represent subclass and part-of relations.

We use the term *sapling* to refer to the tree representing a usually shallow personal hierarchy. A sapling is composed of a root node $r^i$ and its child, or leaf, nodes $\langle l_1^i, ..l_j^i \rangle$. The root node corresponds to a user's collection, and inherits its name, while the leaf nodes correspond to the collection's constituent sets and inherit their names. We assume that hierarchical relations between a root and its children, $r^i \to l_j^i$, specify broader-narrower relations.

On Flickr, users can attach tags only to photos. A sapling's leaf node corresponds to a set of photos, and the tag statistics of the leaf are aggregated from that set's constituent photos. Tag statistics are then propagated from leaf nodes to the parent node. We define a tag statistic of node $x$ as $\tau_x := \{(t_1, f_{t_1}), (t_2, f_{t_2}), \cdots (t_k, f_{t_k})\}$, where $t_k$ and $f_{t_k}$ are *tag* and its frequency respectively. Hence, a root node's tags, $\tau_{r^i}$, are aggregated from all the leaves' tags, $\tau_{l_j^i}$. These tag statistics can also be used as a feature for determining if two nodes are similar (of the same concept).

Any method that aggregates structured social metadata to learn folksonomies has to address a number of challenges. Social metadata is usually very sparse, with each individual user providing just a small amount of evidence, in the form of tags or nodes, for folksonomy learning. Vocabulary noise, due to idiosyncratic naming conventions, misspellings, and the like, is common, and so is ambiguity and synonymy. Moreover, there is structural noise, with users employing varying, and even conflicting, categorization conventions. Varying levels of expertise and expressiveness are also common, with some users creating fine-grained, expressive categorization schemes, and other users coarse-grained, more general categorization schemes [17].

## B. SAP: Incremental Clustering Approach

In a recent work [17], we investigated a relational clustering method that constructs folksonomies from many personal hierarchies in an incremental manner. The folksonomy construction starts with a seed term (which will become a root of the learned folksonomy). Individual saplings whose roots have the same name as the seed are clustered by using some similarity measure, along with a predefined threshold for merging or splitting nodes. At this step, each merged sapling corresponds to a different sense of the seed term. It also assumes that if root nodes are to be merged, their leaves with similar names will also be merged. One of the merged saplings, i.e., a particular sense of the root term, is then selected as the starting point for growing the folksonomy for that concept. Each leaf name is then used to retrieve other saplings whose roots are similar to the name. Subsequently, these saplings are clustered, and one whose root is most similar to the leaf is then attached to the leaf. Structural inconsistencies, such as loops and shortcuts have to be removed if the attachment process creates them. This procedure is done sequentially until the learned folksonomy reaches at a certain depth.

Since the folksonomy has been constructed incrementally from top to bottom, decisions to merge or split saplings at the top of the folksonomy has to be made and fixed before its lower portions can be learned. Consequently, only a small portion of the folksonomy is considered at each integration step, which can lead to a suboptimal structure.

In this paper, we propose a probabilistic framework for folksonomy learning that overcomes the difficulties of existing approaches. Specifically, by considering each concept term as a node, or a data point, within a complex structure, we allow all similar nodes to merge simultaneously. Consequently, a complex structure will appear as nodes are combined. However, since clustering nodes arbitrarily may lead to a structure with some undesired form, e.g., a graph with loops and shortcuts rather than a tree, we propose a method which exploits structural information to guide the clustering procedure to produce a structure with a desired form.

## III. PROBABILISTIC INTEGRATION OF STRUCTURED DATA

A key idea of folksonomy learning through sapling integration is to merge similar nodes from different saplings. Merging similar root nodes expands the width of the learned tree, while merging the leaf of one sapling to the root of another extends its depth. The merging process has two key sub-components: (1) a similarity function that evaluates how similar a pair of nodes is; (2) a procedure that decides if two nodes should or should not be merged, based on their similarity.

Structural information plays an important role in the merging process. Consider the case where two leaf nodes from different saplings are about to be merged. If their parent nodes belong to different clusters, then the merging process will result in a structure which has two paths going to the merged node, i.e., not a tree. But, how should structural information be used? Intuitively, it can be specified within the similarity function used evaluate the decision to merge nodes. For example, similarity between two leaf nodes may contain information about similarity of their parent (and/or sibling) nodes. Therefore, leaf nodes whose parents are not very similar will be less likely to merge. Alternatively, structural information can be specified explicitly through constraints. Such constraints will prevent leaf nodes from being merged if their parents belong to different clusters.

In this section we present a probabilistic framework for distributed inference, and then investigate in detail alternative ways to introduce structural information into the inference process in order to learn deep, bushy trees from many smaller, shallow trees.

### A. Affinity Propagation

As described in the previous section, we need an inference procedure to merge nodes, while exploiting structural information to guide the clustering to order the integrated data in a specific form, a tree in this context. Affinity Propagation (AP) [7] offers a natural framework to incorporate structural information.

AP is a powerful clustering algorithm that identifies a set of exemplar points that well represent all the points in the data set. The exemplars emerge as messages are passed between data points, with each point assigned to an exemplar. AP tries to find the exemplar set which maximizes the net similarity, or the overall similarity between all exemplars and data points assigned to them. Each exemplar and its data points is considered to be a cluster.

We describe AP in terms of a factor graph [12] on binary variables. As recently introduced by Givoni and Frey [8], the model is comprised of a square matrix of binary variables, along with a set of factor nodes imposed on each row and column in the matrix. Following notation of Ref. [8], let $c_{ij}$ be a binary variable indicating whether node $i$ belongs to node $j$ (or, $j$ is an exemplar of $i$). Let $N$ be a number of data points; consequently, the size of the matrix is $N \times N$.

There are two types of constraints that enforce cluster consistency. The first type, $I_i$, which is imposed on the row $i$, indicates that a data point can belong to only one exemplar ($\sum_j c_{ij} = 1$). The second type, $E_j$, which is imposed on the column $j$, indicates that if a point other than $j$ chooses $j$ as its exemplar, then $j$ must be its own exemplar ($c_{jj} = 1$). AP avoids forming exemplars and assigning cluster memberships, which violates these constraints. Particularly, if the configuration at row $i$ violates $I$ constraint, $I_i$ will become $-\infty$, which is not a very optimal configuration (and similarly for $E_j$).

In addition to constraints, a similarity function $S(.)$ indicates how similar a certain node is to its exemplar. If $c_{ij} = 1$, then $S(c_{ij})$ is a similarity between nodes $i$ and $j$; otherwise, $S(c_{ij}) = 0$. $S(c_{jj})$ evaluates "self-similarity," also called "preference", which should be less than the maximum similarity value in order to avoid all singleton points becoming exemplars, since that configuration would yield the highest net similarity. In general, the higher the value of the preference for a particular point, the more likely that point will become an exemplar. In addition, we can set the same self-similarity value to all data points, which indicates that all points are equally likely to become exemplars.

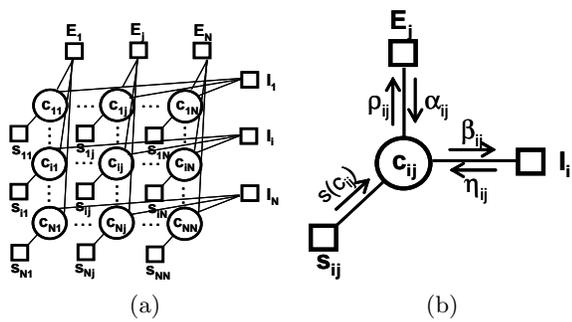

(a)                    (b)

FIG. 2: The original binary variable model for affinity propagation proposed by Givoni and Frey[8]: (a) a matrix of binary hidden variables (circles) and their factors(boxes); (b) incoming and outgoing messages of a hidden variable node from/to its associated factor nodes.

A graphical model of affinity propagation is depicted in Figure 2 in terms of a Factor Graph. In a log-domain, the global objective function, which measures how good the present configuration (a set of exemplars and cluster assignments) is, can be written as a summation of all local factors:

$$\mathbf{S}(c_{11}, \cdots, c_{NN}) = \sum_{i,j} S_{ij}(c_{ij}) + \sum_{i} I_i(c_{i1}, \cdots, c_{iN})$$
$$+ \sum_{j} E_j(c_{1j}, \cdots, c_{1N}). \quad (1)$$

Optimizing this objective function identifies the configuration that maximizes the net similarity $S$, while not violating $I$ and $E$ constraints.

The original work uses max-sum algorithm to optimize this global objective function, and it requires to update and pass five messages as shown in Figure 2(b). Since each hidden node $c_{ij}$ is a binary variable (with two possible values), one can pass a scalar message — the difference between the messages when $c_{ij} = 1$ and $c_{ij} = 0$, instead of carrying two messages at a time. The equations to update these messages are described in greater detail in the Section 2 of Ref. [8].

Once the clustering process terminates, the MAP configuration (exemplars and data points assigned to them) can be recovered as follows. First, we identify an exemplar set by considering the sum of all incoming messages of each $c_{jj}$ (each node in the diagonal of the variable matrix). If the sum is greater than 0 (there is a higher probability that node $j$ is an exemplar), $j$ is an exemplar. Once the set of exemplars $K$ is recovered, each non-exemplar point $i$ is assigned to the exemplar $k$ if the sum of all incoming messages of $c_{ik}$ is the highest compared to the other exemplars.

One can directly apply AP to combine saplings into a more complex structure. In particular, each node in a sapling is treated as a data point. Subsequently, simi-lar data points are grouped together into clusters by AP, while relations between data points are grouped together if their child nodes belong to the same cluster and their parent nodes also belong to the same cluster. Nevertheless, combining saplings in this way could produce an arbitrary graph rather than a tree form. This is because AP does not have an explicit procedure to avoid creating loops and shortcuts.

## B. Expressing Structure through Similarity

Following our previous work [17], we define a similarity measure between nodes in different saplings, which exploits heterogeneous evidence available in the structure of the input data. Basically, the similarity function is a combination of *local similarity* and *structural similarity*. The local similarity between two nodes $i$ and $j$, $localSim(i, j)$, is based on the intrinsic features of $i$ and $j$, such as their tag distributions. The structural similarity, $structSim(i, j)$ is based on features of neighboring nodes. If $i$ is a root of a sapling, its neighboring nodes are all of its children. If $i$ is a leaf node, the neighboring nodes are its parent and siblings. The similarity between nodes $i$ and $j$ is:

$$nodesim(i, j) = (1 - \alpha) \times localSim(i, j) \quad (2)$$
$$+ \alpha \times structSim(i, j),$$

where $0 \leq \alpha \leq 1$ is a weight for adjusting contributions from $localSim(,)$ and $structSim(,)$. To reduce the computational complexity, we assume that nodes with different stemmed names belong to different concepts, and as a result, their similarity is 0. Thus, we only need to evaluate the similarity between a pair of nodes with the same stemmed names to decide whether the nodes refer to the same or different concepts (meanings).

### 1. Local Similarity

To compute $localSim(i, j)$, let $t^{ij}$ be a number of common tags in the top K most frequent tags of nodes $i$ and $j$:

$$localSim(i, j) = min(1, \frac{t^{ij}}{J}), \quad (3)$$

where J is a threshold on a number of common tags.

### 2. Structural Similarity

Structural similarity of two nodes depends on their positions within their saplings. We define three versions: $structSimRR(,)$ which computes structural similarity between two root nodes (root-to-root similarity), $struct-SimLL(,)$, which evaluates structural similarity between a leaf of one sapling to that of another, and $structSimLR(,)$ which evaluates structural similarity between a root of one sapling and the leaf of another (leaf-to-root similarity).

*Root-to-Root similarity*  Two saplings $A$ and $B$ are likely to describe the same concept if their root nodes $r^A$ and $r^B$ have similar names and some of their leaf nodes also have similar names. In this case, there is no need to compute local similarity of these leaf nodes. Structural similarity between two root nodes is then defined as follows:

$$structSimRR(r^A, r^B) \qquad (4)$$
$$= \frac{1}{Z} \sum_{i,j} \delta(name(l_i^A), name(l_j^B)),$$

where $\delta(.,.)$ returns 1 if the both arguments are exactly the same; otherwise, it returns 0; $name(l_i^A)$ is a function that returns the name of a leaf node $l_i^A$ of sapling $A$. $Z$ is a normalizing constant, which is defined as $Z = min(|l^X|, |l^Y|)$, where $|l^X|$ is a number of children of $X$. We use $min(,)$ instead of union. When merging with a relatively small sapling with a larger one, the fraction of common nodes may be very low compared to total number of child nodes. Hence, the normalization coefficient with the union ($Z = union(l^X, l^Y)$), as defined in Jaccard similarity, results in overly penalizing small saplings. $min(,)$, on the other hand, seems to correctly consider the proportion of children of the smaller sapling that overlap with the larger sapling.

*Leaf-to-Leaf similarity*  Two leaf nodes, $l^A$ and $l^B$ are likely to describe the same concept if they have a similar name and some of their siblings also have similar names. Structural similarity between two leaf nodes is defined as follows:

$$structSimLL(l^A, l^B) \qquad (5)$$
$$= \frac{1}{Z-1}((\sum_{i,j} \delta(name(l_i^A), name(l_j^B))) - 1).$$

This is similar to $structSimRR(,)$ but we have to subtract one for excluding the present pair of leaf nodes.

*Root-to-Leaf similarity*  Merging the root of one sapling, $r_B$, with the leaf, $l_A$, of another extends the depth of the learned folksonomy. Since we consider a pair of nodes with different roles, their neighboring nodes also have different roles. This would appear to make them structurally incompatible. Nevertheless, we expect the root of sapling $A$ to share some common features with the root $r_B$. Consequently, we simply define the similarity as,

$$structSimLR(r^B, l^A) = localSim(r^B, r^A). \qquad (6)$$

### 3. Structural Similarity with Cluster Labels

Structural similarity described above does not take the cluster (or concept) of the term into account. For a given pair of terms, we can use cluster labels of their neighboring terms to help decide whether or not they should belong to the same cluster. Intuitively, the more neighboring terms share common cluster labels, the more similar the node pair is. This is along the same line to the earlier work on collective entity resolution [1], where the entity identification decision is based on common neighboring entities rather than references' features.

Let $clust(i)$ be a function which returns the cluster label of node $i$. For the root-to-root structural similarity using cluster labels, we modify Eq. 4 simply by replacing $name()$ with $clust()$. In other words, $structSimRR(r^A, r^B)$ is a normalized intersection between cluster labels of $A$'s leaves and $B$'s leaves.

For the leaf-to-leaf similarity on a pair of leaf nodes, we can only consider the cluster label of their roots rather than all of their siblings. This is because the cluster labels of their root nodes have already taken cluster labels of their siblings into account. Consequently, $structSimLL(l^A, l^B)$ with cluster labels is simply computed from $\delta(clust(r^A), clust(r^B))$.

### 4. Negative Similarity

The structural similarity above does not provide an explicit force to discourage clustering that may cause incoming links from different parent concepts. In some settings, e.g.,[1], the negative similarity can be applied to discourage the merge that violates a constraint. In our context, nevertheless, this similarity is inapplicable since it is imposed on pairwise basis. Specifically, suppose that a root node is formed as a representative (exemplar) of a certain cluster, leaf nodes having different parent clusters can still be legally merged to the cluster. This is due to the fact that AP only considers similarity of nodes to their exemplars. As a result, such configuration does not violate any constraints; therefore, incoming links from different clusters are still permitted.

## C. Expressing Structure through Constraints: Relational Affinity Propagation (RAP)

We extend AP to add structural constraints that will ensure that the learned folksonomy makes sense – no loops, and, to the extent possible, forms a hierarchy. Since we want the learned structure to be a tree, all nodes assigned to some exemplar must have their parent nodes in the same cluster, i.e., assigned to the same exemplar. To achieve this, we must enforce the following two constraints: (1) merging should not create incoming links to a cluster, or concept, from more than one parent cluster (single parent constraint); (2) merging should not create an incoming link to the root of the induced tree (no root parent constraint). For the second constraint, we can simply discard all sapling leaves that are named similar to the tree root. Hence, we only need to enforce the first constraint. The first constraint will be violated if leaf nodes of two saplings are merged, i.e., assigned to the same exemplar, while the root nodes of these saplings are assigned to different exemplars. Consequently, the leaf cluster will have multiple parents pointing to it, which leads to an undesirable configuration.

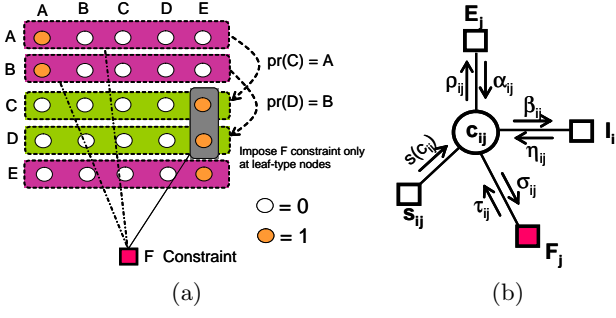

FIG. 3: Relational Affinity Propagation (RAP) proposed in this paper. (a) Schematic diagram of the matrix of binary hidden variables (circles). Variables within the green area correspond to leaf nodes, while those within the pink area correspond to root nodes of saplings. Filled-in circles stand for exemplars. We omit $E$, $I$ and $S$ factors to simplify the diagram. (b) Factor graph representation of RAP.

Let $pa(.)$ be a function that returns the index of the parent node of its argument, and $explr(.)$ be a function that return the index of the argument's exemplar. The factor $F$, "single parent constraint", checks the violation of multiple parent concepts pointing to a given concept. The constraint is formally defined as follows:

$$F_j(c_{1j}, \cdots, c_{Nj}) = \begin{cases} -\infty & \exists i, k : c_{ij} = 1; c_{kj} = 1; \\ & explr(pa(i)) \neq explr(pa(k)), \\ 0 & \text{otherwise.} \end{cases}$$ 
(7)

Figure 3(a) illustrates the way we impose the new constraint on the binary variable matrix. The configuration shown in the figure is valid since both $C$ and $D$ belong to the same exemplar $E$ and their parents, $A$ and $B$, belong to the same exemplar $A$. However, if $c_{BB} = 1$, then the configuration is invalid, because parents of nodes in the cluster of exemplar $E$ will belong to different exemplars ($A$ and $B$). This constraint is imposed only on leaf nodes, because merging root nodes will never lead to multiple parent. The global objective function for RAP is basically Eq. 1 plus $\sum_j F_j(c_{1j}, \cdots, c_{Nj})$. The F-constraint acts as a penalty factor that penalizes the merging that leads to multiple parents. Integrating saplings in such a way that maximizes this objective function will produce a structure with clusters of similar nodes, while all nodes in each cluster must have their parents coming from the same cluster. As in AP, we use max-sum algorithm to optimize this global objective function, which requires passing two additional messages.

To extend AP, we modify the equations for updating the messages $\rho$, $\beta$ and also derive 2 additional messages:

$\sigma$ and $\tau$ to take into account this additional constraint. Following the max-sum message update rule from a variable node to a factor node (cf., eq. 2.4 in Chapter 8 of [2]), the message update formulas for $\rho$, $\beta$ and $\sigma$ are simply:

$$\rho_{ij} = S(i, j) + \eta_{ij} + \tau_{ij},$$ 
(8)

$$\beta_{ij} = S(i, j) + \alpha_{ij} + \tau_{ij},$$ 
(9)

$$\sigma_{ij} = S(i, j) + \alpha_{ij} + \eta_{ij}.$$ 
(10)

For deriving the message update equation for $\tau$, we have to consider two cases: $i = j$ and $i \neq j$, i.e., the $\tau$ message to the nodes on the diagonal and $\tau$ for the rest. For simplicity, we also assume that all leaf nodes have their index numbers less than any roots. Hence, leaf node indices run from 1 to $L$, where $L$ is the number of leaves.

For the case $i = j$ (for the diagonal nodes $c_{jj}$), we have to consider the update message for $\tau$ in two possible settings: $c_{jj} = 1$ and $c_{jj} = 0$ ($\tau_{jj}(1)$ and $\tau_{jj}(0)$ respectively), and then find the best configuration for these settings. The max-sum message update rule from a factor node to a variable node when $c_{jj} = 1$ is [2]:

$$\tau_{jj}(1) = \max_{\mathbf{s}^{\{j\}}} \Big( \sum_{k \in \mathbf{s}^{\{j\}}; k \neq j} \sigma_{kj}(1) + \sum_{l \notin \mathbf{s}^{\{j\}}; l \neq j} \sigma_{lj}(0) \Big).$$ 
(11)

For $c_{jj} = 0$, it is

$$\tau_{jj}(0) = \sum_{k=1:L; k \neq j} \sigma_{kj}(0),$$ 
(12)

where $\mathbf{s}^{\{j\}}$ is a subset of leaf nodes (including $j$) that share the same parent exemplar. Formally, $\mathbf{s}^{\{j\}} \in \mathbf{T}$; $\mathbf{T} \supset \{1, \cdots, L\}$; $\{j\} \subset \mathbf{s}^{\{j\}}$ and all $k$ in $\mathbf{s}^{\{j\}}$ share the same parent exemplar. For Eq. 11, it favors the "valid" configuration (the values of $c_{kj}$), which maximizes the summation of all incoming messages to the factor node $F_j$. The example of a valid configuration in this case is as follows. Suppose we have only 3 leaf nodes: $k$ and $k'$ and $k''$. We would say $< c_{kj} = 1, c_{k'j} = 1, c_{k''j} = 0 >$ is a valid configuration if $k$ and $k'$ have their parents belonging to the same exemplar.

For Eq. 12, since no other nodes can belong to $j$, the valid configuration simply sets all $c_{kj}$ to 0. Note that we omit $F_j$ from the above equations since invalid configurations are not very optimal, so that they will never be chosen. Thus, $F_j$ is always 0.

From Eq. 11 and Eq. 12, the scalar message $\tau_{jj}$ is simply:

$$\tau_{jj} = \tau_{jj}(1) - \tau_{jj}(0) = \max \begin{cases} \max_{\mathbf{s}^{\{j\}}} \sum_{k \in \mathbf{s}^{\{j\}}; k \neq j} \sigma_{kj} \\ 0 \end{cases}$$ 
(13)

For $i \neq j$, we also have to consider the same subcases. For $c_{ij} = 1$, we have:

$$\tau_{ij}(1) = \max_{\mathbf{S}^x} \big( \sum_{k \in \mathbf{S}^x; k \neq i} \sigma_{kj}(1) + \sum_{l \notin \mathbf{S}^x; l \neq i} \sigma_{lj}(0) \big). \quad (14)$$

For $c_{ij} = 0$, we have

$$\tau_{ij}(0) = \max \big( \sum_{k \neq i} \sigma_{kj}(0), \max_{\mathbf{S}} \big( \sum_{k \in \mathbf{S}; k \neq i} \sigma_{kj}(1) \quad (15)$$
$$+ \sum_{l \notin \mathbf{S}; l \neq i} \sigma_{lj}(0) \big) \big),$$

where $\mathbf{S} \in \mathbf{T}$; $\mathbf{T} \supset \{1, \cdots, L\}$, and all $k$ in $\mathbf{S}$ share the same parent exemplar without the restriction that $\mathbf{S}$ must contain $x$. When $j$ is a root node, the leaf node $i$ will never have the multiple-parent conflict with $j$, but we still need to check whether other merging leaf nodes share the same parent exemplar to $i$. Therefore, we set $x = \{j\}$ for this case. Specifically, $\mathbf{S}^x$ in Eq. 14 is replaced by $\mathbf{S}^{\{j\}}$. When $j$ is a leaf node, however, we have to make sure that node $i$, $j$ and other merging leaf nodes have the same parent exemplar. Thus, we set $x = \{i, j\}$. In other words, we substitute $\mathbf{S}^{\{i,j\}}$ for $\mathbf{S}^x$ in Eq. 14. In $c_{ij} = 0$ case, the best configuration may or may not have $j$ as the exemplar, which is different from the $c_{ij} = 1$ case that requires the best configuration necessarily having $j$ as the exemplar.

The scalar message $\tau_{ij}$, which is a difference between $\tau_{ij}(1)$ (Eq. 14) and $\tau_{ij}(0)$ (Eq. 15) is as follows:

$$\tau_{ij} = \min \big( \max_{\mathbf{S}^x} \sum_{k \in \mathbf{S}^x; k \neq i} \sigma_{kj}, \quad (16)$$
$$\big( \max_{\mathbf{S}^x} \sum_{k \in \mathbf{S}^x; k \neq i} \sigma_{kj} - \max_{\mathbf{S}} \sum_{l \notin \mathbf{S}; l \neq i} \sigma_{lj} \big) \big).$$

From the above equation, since the first argument of the formula is always larger than, or equal to the second one, its shorter form is simply:

$$\max_{\mathbf{S}^x} \sum_{k \in \mathbf{S}^x; k \neq i} \sigma_{kj} - \max_{\mathbf{S}} \sum_{l \notin \mathbf{S}; l \neq i} \sigma_{lj}. \quad (17)$$

There is one specific case that the above equation does not cover. The case appears when both $i$ and $j$ are leaf nodes and do not share the same parent exemplars. Therefore, the case $c_{ij} = 1$ should never happen, and that makes $\tau_{ij} \to -\infty$. In other words, we will always prefer $c_{ij} = 0$ to $c_{ij} = 1$. As a result, the scalar message for this case is defined as,

$$\tau_{ij}(explr(pa(i)) \neq explr(pa(j))) = -\infty. \quad (18)$$

For sake of simplifying implementation, we can use any negative value instead of $-\infty$ to simply tell the inference procedure that we always favor $c_{ij} = 0$ in this case.

The inference of exemplars and cluster assignments starts by initializing all messages to zero and keeps updating all messages for all nodes iteratively until convergence. One possible way to determine the convergence is to monitor the stability of the net similarity value, $\sum_{i,j} S_{ij}(c_{ij})$, as in the original AP.

Recovering MAP exemplars and cluster assignments can be done in a slightly different way to the original AP with one extra step, in order to guarantee that the final graph is in a tree form. In particular, for a certain exemplar, we sort its members by their similarity value in descending order. The parent exemplar of a cluster of nodes is determined as follows. If the exemplar of the cluster is a leaf node, the parent exemplar of the cluster is the parent exemplar of the exemplar. Otherwise, the parent exemplar of the highest-ranked leaf node will be chosen. We then split all member nodes that have different parent exemplars to that of the cluster. Note that a more sophisticated approach to this task may be applied: e.g., once split, find the next best valid exemplar to join. However, this more complex procedure is very cumbersome – the decision to re-join a certain cluster may recursively result in the invalidity of other clusters.

Note that RAP can be extended to induce other structure types such as DAG. In DAG case, we simply change the condition in Eq. 7. In particular, for a certain exemplar, its leaf nodes can now have multiple parents, but there will be no descendant nodes of its root nodes belonging to the same exemplar to some ancestor nodes of its leaf nodes.

## Computational Complexity

Both AP and RAP use similarity between pairs of nodes to make cluster decisions. Standard similarity function that only relies on node features can be precomputed at the first iteration, and reused throughout the inference process. On the other hand, class label-based similarity has to be evaluated at every iteration.[22] Therefore, the computational complexity of computing class label-based similarity grows linearly with the number of iterations.

Let $N$ be a number of all nodes (data points) in the data set. Generally, it requires $O(N^2)$ operations to compute all pairwise similarities. Nevertheless, one can apply the blocking idea, e.g., [14], to significantly reduce the number of such pairwise computations. We use a simple blocking scheme, only comparing sapling nodes that share the same stemmed name (we assume that terms having different stemmed names will never get clustered together). Let $M$ be the number of unique stem terms. Hence, for each stem term, there are $\frac{N}{M}$ nodes to be compared on average; as a result, the computational complexity of pairwise similarity reduces to $O((\frac{N}{M})^2)$.

To determine the computational complexity of the clustering procedure, in each iteration AP requires to pass messages to $O((\frac{N}{M})^2)$ nodes. Therefore, the number of operations is proportional to the number of node pairs to be compared. RAP, however, uses additional operations to update $\tau$ messages. Specifically, it needs to (1) update all cluster labels; (2) group nodes that share the same parent. For each node group with the same stem name, the first operation requires sorting nodes by their message values, which can be done in $O(\frac{N}{M}log(\frac{N}{M}))$ operations. The second step can be done in $O(\frac{N}{M})$ operations with a proper data structure. Consequently, RAP requires an additional $O(N(1 + log\frac{N}{M}))$ operations per iteration compared to AP.

## IV. EVALUATION ON REAL-WORLD DATA

We evaluate the different settings described in the previous section on real-world data collected from Flickr and used in recent studies [16, 17]. This data set contains collections and their constituent sets (or collections) created by a subset of Flickr users who are members of seventeen nature and wildlife photography groups. These users had many other common interests, such as travel and sports, arts and crafts, and people and portraiture. All the tags associated with images in the set were also extracted. We stemmed tags, set, and collection names. In all, the data set contains 20,759 saplings created by 7,121 users. A small fraction of these saplings are multi-level. We manually selected 32 seed terms and used the following heuristic to identify relevant saplings. First, we selected saplings whose root names were similar to the seed term. We then used the leaf node names of these saplings to identify other saplings whose root names were similar to these names, and so on, for two iterations.

To compare the different strategies for exploiting structural information, we apply the two clustering procedures, AP and RAP, with different similarity functions, to these data sets. We used the following similarity functions: (1) *local*: only local similarity; and (2) *hybrid*: local and structural similarity; and (3) *class-hybrid*: local and structural similarity using class labels. To make this work comparable to [17], we used the following parameter values in the similarity functions: in local similarity Eq. 3, we set the number of top tags K = 40, and the number of common tags J = 4; in the hybrid similarity function, the weight combination between local and structural similarity is $\alpha = 0.9$ when comparing two nodes that are *both* roots or leaves, and $\alpha = 0.2$ when one node is a root and the other a leaf. Note that unlike [17], there is no need to set the clustering threshold, since exemplars emerge and compete against each other to attract other similar nodes. In all, we have six different settings (two clustering procedures with three similarity schemes).

We apply a strategy similar to [17] to remove inconsistent nodes. Specifically, a inconsistent leaf node is identified by the number of users who specified it, $N_l$, and its parent, $N_r$. If $\frac{N_l}{N_r} < 0.01$, the leaf node term is highly idiosyncratic, and we classify it as inconsistency. Moreover, if there is only one leaf node and a few root nodes in a certain cluster, we will split the leaf node out of the cluster. This heuristic helps to remove concepts that are less relevant to the seed term of the folksonomy.

| Metric | Similarity Scheme | Avg Rank |
|--------|-------------------|----------|
| LR | *local* | 1.71 |
| | *hybrid* | **1.39** |
| | *class* | 1.87 |
| mTO | *local* | **1.55** |
| | *hybrid* | 2.32 |
| | *class* | 1.81 |
| Conflict | *local* | 2.84 |
| | *hybrid* | **1.39** |
| | *class* | 1.68 |
| NetSim | *local* | **1.48** |
| | *hybrid* | 2.45 |
| | *class* | 2.03 |

(a) AP

| Metric | Similarity Scheme | Avg Rank |
|--------|-------------------|----------|
| LR | *local* | 1.61 |
| | *hybrid* | **1.39** |
| | *class* | 1.94 |
| mTO | *local* | **1.61** |
| | *hybrid* | 2.26 |
| | *class* | 1.81 |
| Conflict | *local* | 2.29 |
| | *hybrid* | **1.39** |
| | *class* | 2.10 |
| NetSim | *local* | **1.39** |
| | *hybrid* | 2.35 |
| | *class* | 2.23 |

(b) RAP

TABLE I: The table compares the performance of (a) AP and (b) RAP, when using different similarity schemes on various metrics. The numbers show the average ranks across all 32 seeds. The lower rank, the better performance.

### A. Evaluation Methodology

We measure the performance of the different learning strategies by measuring the properties of the learned tree. Specifically, we evaluate both the *quality* and *structure* of the learned tree (folksonomy). The quality of the learned folksonomy is determined by comparing it to a reference taxonomy. Following methodology described in [17], we use the taxonomy for classifying web pages from the Open Directory Project (ODP)[23] as a reference hierarchy. Since the ODP hierarchy is relatively large, we only consider the portion of it that overlaps the Flickr data set. We apply two metrics: modified Taxonomic Overlap (mTO) [16], and Lexical Recall (LR). Lexical Recall measures term overlap between the learned and reference taxonomies, independent of their structure. mTO measures how well the learned hierarchy preserves parent-child relations found in the reference taxonomy.

For structural evaluation, we apply two metrics: (1) net similarity (*NetSim*); (2) the number of structural conflicts (*Conflicts*). Net similarity measures how well the approach can combine similar smaller structures. It is computed by summing similarities of all nodes to their exemplars. To make all settings comparable, we use Jaccard similarity of the top tags to compute *NetSim*. The number of conflicts measures the structural integrity of the learned tree. It is given by the number of nodes whose parents belong to different clusters. This number is calculated at the end of the final iteration, just before the last step that removes structural conflicts that may still appear. The smaller the value, the more consistent the learned structure.

## B.  Results

We measure how using structural information, either through structural similarity or through structural constraints, affects the quality of the learned folksonomy. To begin, we first evaluate the performance of different similarity schemes (with or without structural information) by running them with AP and RAP. Since all learning stratategies tend to produce more than one tree, we average their performance across all induced trees. We report performance of each learning strategy on a particular metric by ranking it against all other strategies and averaging the rankings across all data sets. This gives a measure of how often a strategy outperforms others. Average rankings are summarized as in Table I(a) for AP and in Table I(b) for RAP.

From Table I, all similarity schemes perform in a similar manner in both AP and RAP. Specifically, structural information in the similarity function (*hybrid* and *class-hybrid*) does help reduce the number of structural conflicts in both AP and RAP. Nevertheless, these similarity functions performed worse on $mTO$ and *NetSim*. This is because they are more stringent than *local*, and cluster fewer saplings together in the folksonomy learning task where individual saplings are rather sparse. Therefore, "similar" structures are less collapsed as indicated by lower *NetSim*. Not surprisingly, these similarity functions do not improve $mTO$ scores over *local*. This is because $mTO$ favors deeper trees to shorter ones if the nodes are ordered correctly. Nevertheless, we hypothesize that in domains where individual structures contain rich information, *hybrid* similarity should outperform *local* similarity.

For *LR*, structural information through *hybrid* similarity can help recover more concepts. This is because learning strategies with similarity function can exploit structural information when local information is not sufficient. However, *class-hybrid* performs worse than *hybrid* in *LR* and in the other metrics. We speculate that class labels at the beginning of the learning process may not be reliable enough, and that leads to the worse performance.

The results of keeping the similarity function fixed, and studying the effectiveness of the clustering strategy are shown in Table II. RAP generally outperforms AP on almost all measures. Specifically, it recovers more concepts (better *LR* score), learns structures better aligned with the reference hierarchy (better $mTO$), produces significantly more consistent structures (fewer *Conflicts*). However, RAP produces trees with lower net similarity (*NetSim*), since it contains more stringent criteria to merge saplings than AP. Note that

Next, we compare RAP with *local* similarity, found to be superior to alternative clustering schemes, to the previous folksonomy learning approach SAP [17]. Unlike SAP, the methods proposed in this paper generally return more than one tree. We simply evaluate the most *popular* tree, which has the largest number of merged nodes at the root level. Figure 4 displays an example of the most

| | Clustering Scheme | |
|---|---|---|
| Measure | AP | RAP |
| LR | 1.35 | 1.35 |
| mTO | 1.42 | **1.29** |
| Conflict | 1.97 | **1.00** |
| NetSim | **1.39** | 1.55 |

(a) *Local* Similarity

| | Clustering Scheme | |
|---|---|---|
| Measure | AP | RAP |
| LR | 1.29 | 1.29 |
| mTO | 1.48 | **1.26** |
| Conflict | 1.97 | **1.00** |
| NetSim | 1.48 | **1.45** |

(b) *Hybrid* Similarity

| | Clustering Scheme | |
|---|---|---|
| Measure | AP | RAP |
| LR | 1.29 | **1.19** |
| mTO | 1.48 | **1.29** |
| Conflict | 1.97 | **1.00** |
| NetSim | **1.32** | 1.58 |

(c) *Class-Hybrid* Similarity

TABLE II: The table compares the performance between AP and RAP when using (a) *local*, (b) *hybrid* and (c) *class-hybrid* similarity on various metrics. The numbers show the average ranks across all 32 seeds. The lower rank, the better performance.

*popular* tree of `bird`, which is induced by RAP with *local* similarity.

Due to space limitations, we only report the quality of the learned folksonomy, as measured by $mTO$ scores and the number of overlapping paths (#OPaths) to the reference hiearchy. For #OPaths, we consider two paths are "overlapping" if their root (source) nodes share the same name; and their leaf (sink) nodes share the same name. Therefore, the number of overlapping paths are enumerated by counting how many leaves in the learned folksonomy share similar names to some leaves in the reference hierarchy. Since $mTO$ is computed from the overlapping paths, the approach that yields higher $mTO$ and higher #OPaths at the same time is preferable. Note that we cannot compare RAP with SAP on *Conflicts* and *NetSim* metrics because of their algorithmic difference. In particular, RAP provides an approximate solution, which often contains some conflicts in a "difficult" case. In SAP, however, conflicts are heuristically removed as a tree grows. Moreover, it's impossible to compute *NetSim* on a SAP tree since SAP does not identify any exemplars.

As shown in Table III, RAP with *local* similarity can produce more consistent taxonomies compared to SAP (15 vs. 12 cases). Moreover, if considering both numbers of comparable paths (#OPaths) and $mTO$, RAP+*local* is clearly superior to SAP (14 vs. 4 cases). Specifically, the former produces more consistent structures on a higher number of comparable paths, with respect to the reference hierarchy.

Nevertheless, $AUT$ (a metric for measuring how detailed a folksonomy from its bushiness and depth [17])

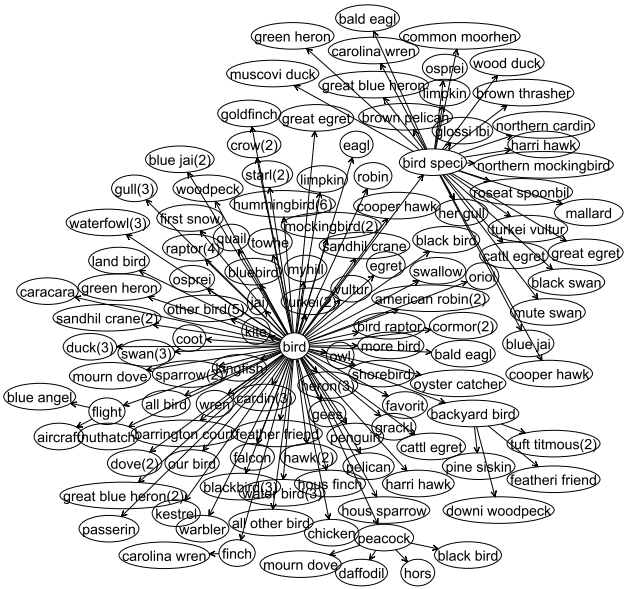

FIG. 4: A folksonomy learned for **bird** using RAP with *local* similarity. Due to space limiations, concepts with a similar name that do not merge together are visualized in a single node with a number in parentheses that enumerates their number.

and *LR* scores (not presented in the table) of trees produced by RAP vs SAP (6 vs. 24 cases, and 12 vs. 19 cases respectively). This is because of the nature of AP and its extension, RAP, that allows different trees to emerge simultaneously. In many cases, these trees attract the most similar structures to it. Compared to SAP, which greedily grows one tree at a time, attracting all similar concepts to it, RAP assigns concepts to different trees with which they have the best fit. Since we only consider one of the trees in the evaluation, there is a high chance that the selected tree contains relatively fewer unique concepts and so is not bushier than SAP's tree.

The overall experimental results clearly suggest that the proposed approach (RAP), which incorporates structural information through constraints during probabilistic inference process can learn better, more consistent structures. We speculate that RAP can be even more advantageous in domains where heuristics for correcting the learned structure to a specific form are difficult to specify and expensive to carry out.

## V. RELATED WORK

Learning from structured data has emerged as a popular research area in machine learning. Closest to ours is work on learning systems of concepts [11] and learning hierarchical topics from words in documents [3]. Nevertheless, our work is fundamentally different from them in that, we "align" or integrate many small shallow hi-erarchies which are *explicitly* specified by users. Consequently, our work attempts to find the best "alignment" or "integration," which maximizes the similarity between concepts and has no structural inconsistencies.

| | SAP | | RAP + *local* | |
|---|---|---|---|---|
| seeds | #OPaths | mTO | #OPaths | mTO |
| africa | 27 | **0.895** | 37 | 0.869 |
| anim | 92 | **0.659** | 106 | 0.656 |
| asia | **85** | 0.788 | 43 | 0.785 |
| australia | 27 | 0.665 | 46 | **0.672** |
| bird | 22 | **0.755** | 38 | 0.714 |
| build | 0 | 0.000 | 0 | 0.000 |
| canada | 27 | 0.587 | 47 | **0.689** |
| cat | 0 | 0.000 | 1 | **0.508** |
| c. america | 2 | 0.754 | 6 | **0.863** |
| citi | 0 | 0.000 | 0 | 0.000 |
| countri | **4** | **0.665** | 1 | 0.000 |
| craft | 0 | 0.000 | 14 | **0.400** |
| dog | 1 | 1.000 | 4 | 1.000 |
| europ | **301** | **0.670** | 133 | 0.596 |
| fauna | **31** | 0.490 | 14 | **0.529** |
| fish | 0 | 0.000 | 7 | **0.672** |
| flora | 18 | 0.481 | 28 | **0.512** |
| flower | 1 | **1.000** | 9 | 0.783 |
| insect | 5 | **0.924** | 18 | 0.836 |
| invertebr | 1 | **1.000** | 26 | 0.752 |
| n. america | 118 | 0.576 | **182** | **0.683** |
| plant | 7 | 0.735 | 11 | **0.795** |
| reptil | 3 | 0.622 | **4** | **0.625** |
| s. africa | 3 | 0.600 | **4** | 0.600 |
| s. america | 15 | **0.832** | 28 | 0.637 |
| sport | 27 | 0.647 | **114** | **0.649** |
| u. kingdom | 82 | **0.724** | 135 | 0.620 |
| u. state | 55 | 0.749 | **133** | **0.823** |
| urban | 0 | 0.000 | 4 | **0.603** |
| vertebr | 0 | 0.000 | 3 | **1.000** |
| world | **475** | **0.461** | 44 | 0.432 |
| Summary | **1429**(sum) | 0.557(avg) | 1240(sum) | **0.629**(avg) |

TABLE III: The table compares the performance on *mTO* of the proposed approach,RAP with *local* similarity scheme, to the previous work, SAP [17]. The table also reports a number of comparable paths,*#OPaths* to the reference hierarchies.

In the social web domain, most of the previous work utilizes tag statistics as evidence for learning broader/narrower relations between concepts [15, 19]. Since these works are based on tag statistics, they are likely to suffer from the "popularity vs generality" problem, where a tag may be used more frequently not because it is more general, but because it is more popular among users. Moreover, the approaches only focus on learning pair-wise relations rather than constructing full hierarchies. These are all different from the present work, which focuses on exploiting existing relations and combining them together into full hierarchies.

Folksonomy integration is similar to ontology alignment [6, 20] in that both identify matches between con-cepts in pairs of structures. Nevertheless, ontology alignment differs from our problem, since in ontology alignment there are typically just a few structures to align, and those structures are deep and semantically rich. Here, we focus on the much noisier setting, where there are many smaller fragments created by end users with a variety of purposes in mind. A recent work [17] addressed this problem by applying the relational clustering approach to exploit structure and tag statistics to incrementally attach relevant saplings to the learned folksonomies. In that work, since the folksonomy has been constructed incrementally from top to bottom, only a small portion of the folksonomy is considered at each integration step, which may lead to a sub-optimal structure. This is different from the method described in this paper that integrates all fragments simultaneously into a unified tree.

Affinity propagation, on which the present work is based, has been applied to many clustering problems, e.g. segmentation in computer visions [13], because it provides a natural way to incorporate constraints while simultaneously improving the net similarity of the cluster assignments, which is not trivial to handle with standard clustering techniques. In addition, no strong assumption is required on the threshold, which determines whether clusters should be merged or not. Moreover, the cluster assignments can be changed during the inference process as suggested by the emergence of exemplars, compared to "incremental" clustering approaches (e.g., [1]), in which previous clustering decisions cannot be changed. To the best of our knowledge, ours is the first extension of AP algorithm that can learn tree structures from many sparse and shallow trees.

Several other statistical relational learning (SRL) approaches may be applicable to this class of problems. For example, Markov Logic Networks (MLN) [18] and Probabilistic Similarity Logic (PSL) [4], are generic frameworks for solving probabilistic inference problems. They may be used for folksonomy learning by translating similarity function as well as constraints into logical predicates. Since our similarity function is continuous, hybrid MLN (HMLN) [21] would be required. Nevertheless, AP framework is more preferable for the present problem due to its simplicity. For some problems which require to model multiple types of relations and constraints, MLN and PSL may be more suitable.

## VI. DISCUSSION AND CONCLUSION

We described a probabilistic approach, RAP, that extends the distributed inference approach used by affinity propagation to combine a large number small structures into a few, integrated complex structures. We studied two different ways to incorporate structural information into the inference process, and applied the approach to the folksonomy learning problem. The experimental results suggest that, in folksonomy learning setting, the approach that incorporates structural information through constraints, RAP, can help produce high quality folksonomies, often better than those learned by the current state-of-the-art approach. In addition, the proposed approach is general enough for other domains, in which partial structures are specified, such as tags bundles in *Delicious*, files and folders in personal workspaces and semantic networks.

Regarding future work, we would like to extend the approach to induce other classes of structures, e.g., DAGs. We would also like to extend RAP to apply to other structure learning problems, such as alignment of biological data. Finally, we would like incorporate more efficient inference algorithm and compare the aproach to other statistical relational learning (SRL) approaches.

### Acknowledgements

This material is based upon work supported by the National Science Foundation under Grant No. IIS-0812677.

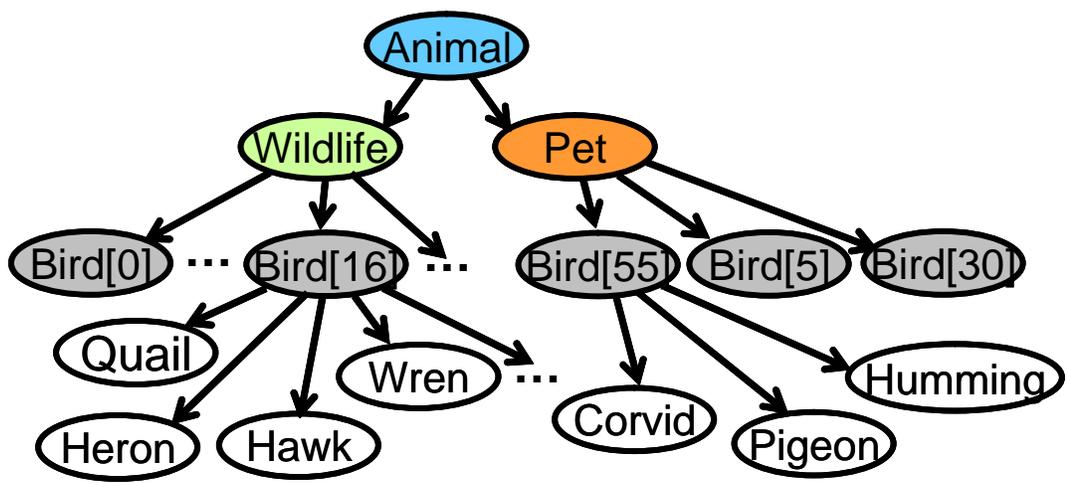

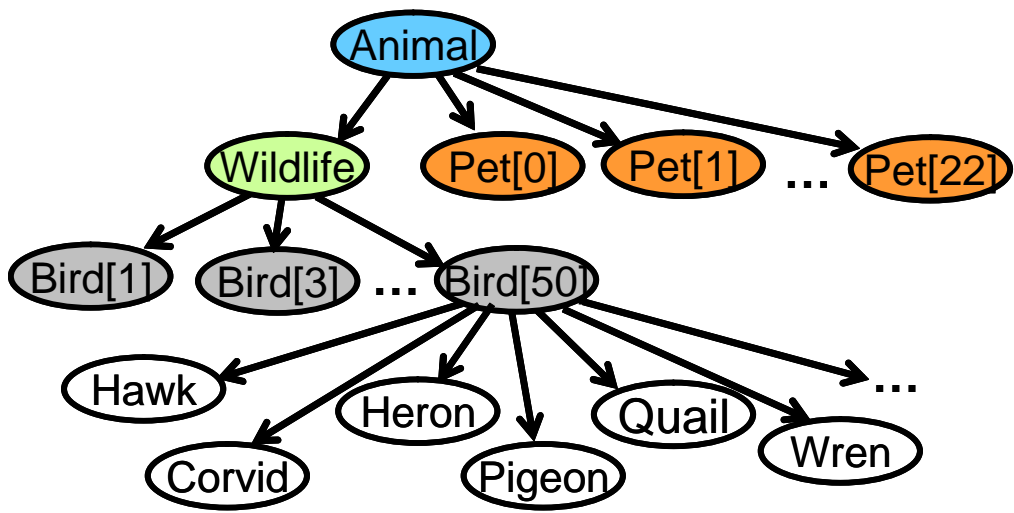